\documentclass[review,hidelinks,onefignum,onetabnum]{siamart220329}

\usepackage{graphicx,amsmath,amssymb,txfonts}
\usepackage{caption}
\usepackage{tabularx}
\usepackage{enumitem}
\usepackage{subfigure,color}
\usepackage{hyperref}
\usepackage{booktabs}
\usepackage{threeparttable}
\usepackage{makecell}
\usepackage{epstopdf}
\usepackage{float}
\usepackage{appendix}
\usepackage{tikz}
\usepackage{cases}
\usepackage[amsmath,thmmarks]{ntheorem}
\ifpdf
  \DeclareGraphicsExtensions{.eps,.pdf,.png,.jpg}
\else
  \DeclareGraphicsExtensions{.eps}
\fi

{
\newtheorem{thm}{\textbf{Theorem}}[section]

\newtheorem{coro}[thm]{\textbf{Corollary}}

\newtheorem{lem}[thm]{\textbf{Lemma}}

\newtheorem{assu}[thm]{\textbf{Assumption}}
\newtheorem{rem}[thm]{\textbf{Remark}}

}

\newtheorem*{pro}{\textbf{Proof}}

\headers{NN-HiSD for Solution Landscape}{Y. Liu, L. Zhang, J. Zhao}

\title{Neural Network-based High-index Saddle Dynamics Method for Searching Saddle Points and Solution Landscape\thanks{Submitted to the editors DATE.
\funding{This work was supported by the National Natural Science Foundation of China (No.12225102, T2321001, 12288101 and 12301520) and Beijing Outstanding Young Scientist Program (JWZQ20240101027).}}}

\author{Yuankai Liu\thanks{School of Mathematical Sciences, Peking University, Beijing, 100871, P.R. China. 
  (\email{liuyk@pku.edu.cn}).}
\and Lei Zhang\thanks{Beijing International Center for Mathematical Research, Center for Quantitative Biology, Center for Machine Learning Research, Peking University, Beijing, 100871, China. (\email{zhangl@math.pku.edu.cn}).}
\and Jin Zhao\thanks{Academy for Multidisciplinary Studies, Capital Normal University, and Beijing National Center for Applied Mathematics, Beijing, 100048, China.
  (\email{zjin@cnu.edu.cn}).}
  }

\usepackage{amsopn}

\ifpdf
\hypersetup{
  pdftitle={Neural Network-based High-index Saddle Dynamics Method for Searching Saddle Points and Solution Landscape},
  pdfauthor={Y. Liu, L. Zhang, J. Zhao}
}
\fi

\begin{document}

\maketitle

\begin{abstract}
The high-index saddle dynamics (HiSD) method is a powerful approach for computing saddle points and solution landscape. However, its practical applicability is constrained by the need for the explicit energy function expression. To overcome this challenge, we propose a neural network-based high-index saddle dynamics (NN-HiSD) method. It utilizes neural network-based surrogate model to approximates the energy function, allowing the use of the HiSD method in the cases where the energy function is either unavailable or computationally expensive. We further enhance the efficiency of the NN-HiSD method by incorporating momentum acceleration techniques, specifically Nesterov's acceleration and the heavy-ball method. We also provide a rigorous convergence analysis of the NN-HiSD method. We conduct numerical experiments on systems with and without explicit energy functions, specifically including the alanine dipeptide model and bacterial ribosomal assembly intermediates for the latter, demonstrating the effectiveness and reliability of the proposed method.
\end{abstract}



\section{Introduction}
The investigation of saddle points on energy landscapes is a central pursuit in computational science, reflecting a critical juncture in disciplines such as physics \cite{PhysRevE.84.025702}, chemistry \cite{ E2010TransitionpathTA,doi:10.1007/BF00532206}, and computer science \cite{Fukumizu2019SemiflatMA,jin2018accelerated}. They offer invaluable insights into the complex behaviors, including phase transitions \cite{PhysRevLett.104.148301, doi:10.1016/j.jcp.2019.06.028, doi:10.1126/science.1253810,doi:10.1016/j.actamat.2009.10.041,PhysRevLett.98.265703}, protein folding \cite{doi:10.1038/nsb0497-305}, and deep neural networks \cite{daneshmand2018escaping, Dauphin2014IdentifyingAA}. 


In general, the computation of high-index (index$>$1) saddle points is more difficult than the stable minima as it has multiple unstable eigen-directions. There exist various numerical algorithms of finding saddle points and minimum energy paths \cite{E2011gentlest,Gould2016dimertype,Li2001minimax}. Among them, the high-index saddle dynamics (HiSD) method serves as an efficient tool in computing any-index saddle points \cite{doi:10.1137/19M1253356}. The HiSD method is then combined with the downward/upward search algorithms \cite{PhysRevLett.124.090601,Yin2020} to construct the solution landscapes of both gradient systems and non-gradient systems \cite{doi:10.1098/rspa.2021.0458,doi:10.1088/1361-6544/abc5d4,doi:10.1137/22M1487965,Shi2022Nematic}. It has garnered increasing attention due to its successful applications in various fields, such as revealing the defect landscape of nematic liquid crystals \cite{doi:10.1088/1361-6544/acc62d,doi:10.1017/S0962492921000088}, the nucleation of quasicrystals \cite{Yin_2021,zhou2024nucleation}, and the excited states of rotational Bose-Einstein condensates \cite{doi:10.1016/j.xinn.2023.100546}. 

However, the HiSD method has relied heavily on the explicit formulations of energy functions, which are often either unavailable or too complex to compute using conventional analytical or numerical methods.
Thus, in situations where the exact form of the energy function is unknown but related data are available, the contribution of the surrogate model becomes crucial. Notably, this importance remains substantial even if the energy function takes a parameterized form. Furthermore, for computationally expensive energy functions, a practical and efficient strategy involves storing the data in advance and then training more computationally efficient surrogate models. 
Deep neural networks (DNNs), grounded in the universal approximation theorem \cite{Cybenko1989,HORNIK1991251,Hornik1989}, have emerged as a powerful tool capable of approximating any given function to an arbitrary degree of accuracy. This inherent capability has made DNNs an attractive choice for surrogate models. 

In this paper, we introduce a novel framework for neural network-based high-index saddle dynamics (NN-HiSD), which utilizes a neural network surrogate model to approximate the energy function and employs HiSD on this surrogate model to compute saddle points. 
Further, the introduction of an automatic differentiation feature \cite{10.5555/3122009.3242010} enhances the model's precision, enabling a discrepancy-correction mechanism that guides the surrogate model toward increasingly accurate representations of the energy functions. To further enhance computational efficiency, we integrate momentum-accelerated optimization techniques, specifically Nesterov's acceleration \cite{Nesterov1983AMF} and the heavy-ball method \cite{doi:10.1016/0041-5553(64)90137-5}, to expedite the convergence of NN-HiSD. 
Based on the universal approximation theorem \cite{Cybenko1989}, we demonstrate that, with sufficient training, the saddle points of the surrogate model are closely aligned with those of the original function. Additionally, we provide a rigorous convergence analysis of the NN-HiSD method. 

To test the effectiveness of the proposed method, we conduct extensive numerical examples with and without explicit energy expressions, specifically focusing on data-driven cases including the alanine dipeptide model and bacterial ribosomal assembly intermediates. We successfully apply the NN-HiSD method to calculate numerous saddle points and construct the solution landscape across various examples.


This paper is organized as follows. In Section 2 we introduce the neural network-based high-index saddle dynamics method. Section 3 focuses on the solution landscape and high-index saddle dynamics method with momentum acceleration. In Section 4 we give the convergence analysis of neural network-based high-index saddle dynamics method. We conduct lots of numerical experiments to validate our method in Section 5. Conclusions are drawn in the last section. 

\section{Neural Network-based High-index Saddle Dynamics Method}
Before we introduce the NN-HiSD method, we first review the classical HiSD method \cite{doi:10.1137/19M1253356}. Given a twice differentiable energy function $E(x)$, we can present the index-$k$ saddle dynamics for calculating an index-$k$ saddle point as follows.
\begin{subnumcases}{\label{2.1}}
    \dot{x} = \beta\left(I-\sum_{i=1}^k 2 v_i v_i^{\top}\right) F(x), \label{2.1a} \\
    \dot{v}_i = -\gamma\left(I-v_i v_i^{\top}-\sum_{j=1}^{i-1} 2 v_j v_j^{\top}\right) G(x) v_i, \quad i=1,2, \ldots, k. \label{2.1b}
\end{subnumcases}
Here $I$ is the identity matrix, $F(x)=-\nabla E(x)$ is the natural force of the energy $E(x)$, $G(x)=\nabla^2 E(x)$ is the corresponding Hessian, $v_1, \cdots, v_k$ are $k$ eigenvectors corresponding to the first $k$ negative eigenvalues of  $G(x)$, and $\beta,~\gamma$ are two positive relaxation parameters.

In this context, the Morse index of a nondegenerate critical point $x^*$ is defined as the number of negative eigenvalues of the Hessian matrix at $x^*$, which equals to the dimensionality of the largest subspace where the associated quadratic form is negative definite \cite{Milnor+1963}. Specifically, an index-1 saddle point, characterized by having exactly one negative eigenvalue, is referred to the transition state that connects two minima on the energy landscape.

In practice, the dimer method in \cite{doi:10.1137/19M1253356} is used to 
approximate the multiplication of the Hessian and the vector $G(x)v_i$ in \eqref{2.1b}: 
$$
H(x,v_i,l)=\frac{F(x-lv_i)-F(x+lv_i)}{2l}.
$$

Indeed, a more crucial task lies in identifying the eigenvectors, $v_i~(i=1,\cdots,k)$. Once these have been determined to a certain level of accuracy by some numerical method, the process of solving for the evolution of $x$ in \eqref{2.1a} is effectively completed. Thus, the discrete form of index-$k$ saddle dynamics \eqref{2.1} is given by
\begin{equation}\label{2.2}
\left\{\begin{array}{l}
x^{(n+1)}=x^{(n)}+\beta_n\left(I-2 \displaystyle\sum_{i=1}^k v_i^{(n)} v_i^{(n)^{\top}}\right) F(x^{(n)}), \\
\left\{v_i^{(n+1)}\right\}_{i=1}^k=\text { EigenSol }\left\{G\left(x^{(n+1)}\right),\left\{v_i^{(n)}\right\}_{i=1}^k\right\}.
\end{array}\right.
\end{equation}
Here $\beta_n$ differs from the one in equation \eqref{2.1a}, as it incorporates $\Delta t$ to simplify the expression. 
``EigenSol'' represents some eigenvector solver with initial values $\big\{\hat v_i^{(n)}\big\}_{i=1}^k$ to compute eigenvectors corresponding to the first $k$ smallest eigenvalues of $\nabla^2 E(x^{(n+1)})$, such as the simultaneous Rayleigh-quotient iterative minimization method (SIRQIT) \cite{longsine1980simultaneous} and locally optimal block preconditioned conjugate gradient (LOBPCG) method \cite{knyazev2001toward}.

We now introduce the NN-HiSD method. The framework of the NN-HiSD method is represented in Fig. \ref{fig1}:
\begin{figure}[htbp]
    \centering
    \includegraphics[width=0.95\textwidth]{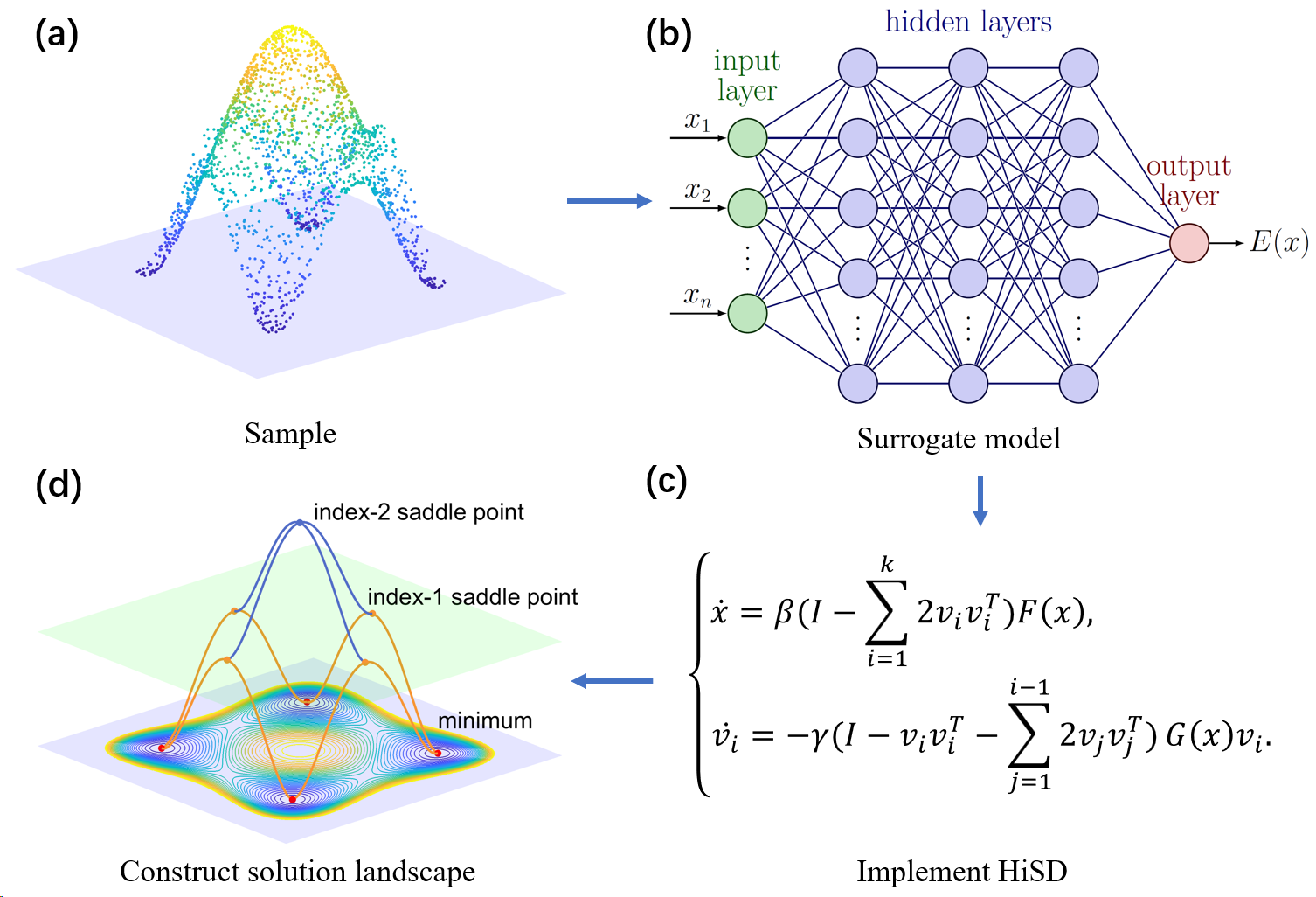}
    \caption{Overview of the framework of the NN-HiSD method. (a) Function values at only a limited set of specific points available. (b) Construct neural network-based surrogate model. (c) Implement the HiSD method with the momentum acceleration. (d) Construct the solution landscape using downward and upward search.}
    \label{fig1}
\end{figure}

The NN-HiSD method is mainly used to address systems for which the explicit expressions for $E(x)$ or the corresponding natural force $F(x)$ in equation \eqref{2.1} are elusive or challenging to derive.  This indicates that we possess function values at only a limited set of specific points, while it remains necessary to compute the corresponding saddle points. 

DNNs are employed as surrogate models in this paper. 
We use a neural network
$
E_{NN}(x;\theta)
$
to approach $E(x)$ with the network's parameters $\theta$. The related loss function is defined as
\begin{equation}\label{3.1}
    \mathcal{L}_1=\left\|E_{NN}(x;\theta)-E(x)\right\|,
\end{equation}
where $\|\cdot\|$ denotes the $L_2$ norm.

For problems with available partial gradient information, this can similarly be integrated into our framework. We can define another loss function as follows:
$$
  \mathcal{L}_2 = \left\| AD_x(E_{NN}(x; \theta)) - \nabla E(x) \right\|, 
$$
where $ AD $ represents Automatic Differentiation \cite{10.5555/3122009.3242010}, a crucial technique in neural networks, and $ AD_x $ denotes the application of AD with respect to the variable $ x $. Therefore, combining \eqref{3.1} we can train the neural networks with the total loss
$$
\mathcal{L}=\mathcal{L}_1+\lambda_2\mathcal{L}_2,
$$
where $\lambda_2$ is a penalty coefficient. 

In many scenarios, we may deal with a class of functions rather than a single function, which are typically represented in parametric form. Specifically, if the precise energy function is given by
$$
E = E(x; \alpha),
$$
where $\alpha$ is a parameter, then the pair $(x, \alpha)$ should be utilized as the input for the neural network. Within this framework, the surrogate model is formulated as $E_{NN}(x, \alpha; \theta)$, where $\theta$ represents the neural network parameters. Importantly, $\alpha$ serves as one of the input variables for the neural network.

For complex systems, the HiSD method may require numerous iterations, which could lead to a slow computation of saddle points. Therefore, it is essential to accelerate the HiSD method. In this paper momentum-accelerated methods are utilized. Specifically, the HiSD method with heavy-ball acceleration ($\text{HiSD}_{\text{Hb}}$) \cite{luo2023accelerated} is defined as below:
\begin{equation}\label{2.3}
\left\{\begin{array}{l}
x^{(n+1)}=x^{(n)}+\beta_n\left(I-2 \displaystyle\sum_{i=1}^k v_i^{(n)} v_i^{(n)^{\top}}\right) F\left(x^{(n)}\right)+\gamma\left(x^{(n)}-x^{(n-1)}\right), \\
\left\{v_i^{(n+1)}\right\}_{i=1}^k=\text { EigenSol }\left\{G\left(x^{(n+1)}\right),\left\{v_i^{(n)}\right\}_{i=1}^k\right\}.
\end{array}\right.
\end{equation}
Here $\gamma \in [0,1)$ is the coefficient of the momentum. And the HiSD method with Nesterov's acceleration ($\text{HiSD}_{\text{NA}}$) reads as
\begin{equation}\label{2.4}
\left\{\begin{array}{l}
w^{(n)}=x^n+\gamma_n\left(x^{(n)}-x^{(n-1)}\right), \\
x^{(n+1)}=w^{(n)}+\beta_n\left(I-2 \displaystyle\sum_{i=1}^k v_i^{(n)} v_i^{(n)^{\top}}\right) F\left(w^{(n)}\right), \\
\left\{v_i^{(n+1)}\right\}_{i=1}^k=\text { EigenSol }\left\{G\left(x^{(n+1)}\right),\left\{v_i^{(n)}\right\}_{i=1}^k\right\} .
\end{array}\right.
\end{equation}
Here $\gamma_n$ is a parameter with several possible options. To be specific, there are two choices available:
\begin{equation*}
  \begin{aligned}
    \text{Choice~ 1}: \gamma_n=&\frac{n}{n+3};\\
    \text{Choice~ 2}: \gamma_n=&\theta_{n+1}^{-1}\left(\theta_n-1\right)~\text{with}~ \theta_{n+1}=\frac{1+\sqrt{1+4 \theta_n^2}}{2},~ \theta_0=1.
  \end{aligned}
\end{equation*}
In our specific experiments, we find that the two choices yield almost identical results. Therefore, for the experiments presented in this paper, we exclusively use the first choice. To achieve better results, we also employ the fixed restart scheme proposed in \cite{WOS:000354710400004}.

We can use the NN-HiSD method to a construct the solution landscape \cite{Yin2020}, which is a pathway map \cite{PhysRevLett.124.090601} consisting of all stationary states, including saddle points and stable minima, and the connections between these states. This procedure includes a downward search algorithm and an upward search algorithm.

The downward search is the main process for searching stationary points starting from a high-index saddle. Given an index-$m$ saddle point $\hat{x}$, the vectors $\hat{v}_1, \ldots, \hat{v}_m$ are the $m$ orthonormal eigenvectors of $\nabla^2 E(\hat{x})$, corresponding to the smallest $m$ eigenvalues $\hat{\lambda}_1< \ldots < \hat{\lambda}_m<0$. We choose a direction $\hat{v}_i$ from the unstable directions $\{\hat{v}_1, \ldots, \hat{v}_m\}$ and slightly perturb the parent state $\hat{x}$ along this direction. The HiSD searching for an index-$k$ ($k<m$) critical point is started from the point $\hat{x}\pm\varepsilon\hat{v}_i$ and the $k$ initial directions from the unstable directions need to exclude $\hat{v}_i$. A typical choice for the initial position and vectors in a downward search is $(\hat{x} \pm \varepsilon\hat{v}_{k+1}, \hat{v}_1, \ldots, \hat{v}_k)$. An arrow is drawn from $\hat{x}$ to this critical point in the solution landscape to illustrate this relationship. This procedure is repeated to newly-found saddle points until no more critical points are identified. 

Conversely, when the parent state is unknown or multiple parent states exist, an upward search algorithm is employed to search for high-index saddles starting from a low-index saddle point or a minimum. Given an index-\(m\) saddle point \(\hat{x}\), more directions \(\hat{v}_1, \ldots, \hat{v}_K\) must be computed at \(\hat{x}\) for the upward search, where \(K > m\) represents the highest index of the saddle being sought. In the HiSD method for finding an index-\(k\) (\(k > m\)) saddle point, we select a direction \(\hat{v}_i\) from the stable directions $\{\hat{v}_{m+1}, \ldots, \hat{v}_k\}$ and initiate the search from \(\hat{x} \pm \varepsilon \hat{v}_i\), where the \(k\) initial directions should include \(\hat{v}_i\). A typical choice for the initial position and directions in an upward search is $(\hat{x} \pm \varepsilon \hat{v}_k, \hat{v}_1, \ldots, \hat{v}_k)$.

\section{Convergence Analysis of Neural Network-based High-index Saddle Dynamics Method}
In this section, we establish that when the approximation of the surrogate model is sufficiently precise, the saddle points of the surrogate model converge closely to those of the original energy function. Furthermore, we prove the local convergence of the NN-HiSD algorithm based on this result. 
Here our analysis is based on the following assumptions:
\begin{assu}
\label{assu:4.1}
The variable $x^*$ represents an index-$k$ saddle point of the potential function $E(x)$.
The initial position $x^{(0)}$ locates in a neighborhood of the saddle point $x^*$,
i.e. $x^{(0)} \in U(x^{*}, \delta) = \{x\mid\|x-x^*\|_2 < \delta\}$ for some $\delta > 0$ such that
\begin{enumerate}[label=\itshape(\roman*)]
	\item There exists a constant $M > 0$ such that
	$\|\nabla^2 E(x) - \nabla^2 E(y)\|_2\leq M\|x-y\|_2$ for all $x,y\in U(x^*,\delta)$;
	\item For any $x \in U(x^*,\delta)$, the eigenvalues $\{\lambda_i\}_{i=1}^d$ of $\nabla^2 E(x)$
	satisfy $\lambda_1 \leq \cdots \leq \lambda_k < 0 < \lambda_{k+1}\leq \cdots \leq \lambda_d$ and
	there exist positive constants $0<\mu<L$ such that $|\lambda_i|\in [\mu, L]$ for $1\leq i \leq d$.
\end{enumerate}
\end{assu}

\begin{assu}	
\label{assu:4.2}
We assume that $v_i^{(n)}$ in the HiSD method \eqref{2.2} satisfies 
$$
\|v_i^{(n)}\|_2 = 1\quad \text{and}\quad \nabla^2 E(x^{(n)}) v_i^{(n)} = \lambda_i^{(n)} v_i^{(n)}, \quad i = 1, 2, \ldots, k.
$$
\end{assu}
\begin{assu}
\label{assu:4.4}
Our surrogate model can be written as
\begin{equation*}
	E_{NN} = E + E_\delta
\end{equation*}
and for all $x,y\in U(x^*,\delta)$, $E_\delta$ satisfies
\begin{enumerate}[label=\itshape(\roman*)]
\item $\|\nabla^2 E_\delta(x) - \nabla^2 E_\delta(y)\|_2 \leq \varepsilon \|x - y\|_2$;
\item $\|\nabla^2 E_\delta(x)\|_2 \leq \varepsilon$;
\item $\|\nabla E_\delta(x)\|_2 \leq \varepsilon$.
\end{enumerate}
\end{assu}
In fact, this assumption is easily satisfied. For this, we give the following remark.
\begin{rem}
To ensure that the assumption holds, it is requisite for the neural network to possess the capability to approximate the original function with arbitrary precision under the $C^3$ norm. For the activation function, we employ the smooth $ \tanh $ function. Drawing upon the foundational principles of the universal approximation theorem \cite{Cybenko1989,HORNIK1991251,Hornik1989}, coupled with an adequately extensive dataset, a sufficiently intricate architectural framework, and thorough training, we can assert with confidence that the assumption is robustly maintained.
\end{rem}
The following inference can immediately be obtained:
\begin{coro}
	\label{coro:4.5}
	Suppose $\varepsilon<\mu$, then $\nabla^2 E_{NN}(x)$ and $\nabla^2 E(x)$ have the same index and
	\begin{equation}
		\|\nabla^2 E_{NN}(x)\|_2 \in [\mu-\varepsilon, L+\varepsilon].
	\end{equation}
\end{coro}

To propose the convergence analysis of NN-HiSD, 
we need to reference the convergence analysis of HiSD as outlined below:
\begin{lem} (\textbf{Theorem 5.2} in \cite{doi:10.1137/22M1487965})
\label{lem:4.3}
Under Assumptions \ref{assu:4.1} and \ref{assu:4.2}, if the initial point $x^{(0)}$ satisfies
\begin{equation*}
	r_0=\|x^{(0)}-x^*\|_2<\min\{\delta, \hat{r}\}, \quad \hat{r}=\frac{2\mu}{M}
\end{equation*}
and $\beta_n=\frac{2}{L+\mu}$ for any $n\geq0$, $x^{(n)}$ converges to $x^*$ as $n\to\infty$ in HiSD with the estimate on the convergence rate
\begin{equation}
	r_n=\|x^{(n)}-x^*\|_2\leq (1-\frac{2}{\kappa+3})^n\frac{\hat{r}r_0}{\hat{r}-r_0}, \quad \kappa=\frac{L}{\mu}.
\end{equation}
\end{lem}

Next, we note that when the surrogate model is sufficiently well-trained, its saddle point will be situated in close proximity to the saddle point of the original function. 
\begin{thm}
\label{thm:4.6}
Suppose that $\varepsilon$ is sufficiently small, precisely $\varepsilon$ satisfies
\begin{equation}
	\varepsilon \leq \min\{\frac{\mu}{2}, \frac{\mu^2}{32M}, \frac{\mu\delta}{12}, M\},
\end{equation}
there exists $x_{NN}$ which satisfies $\nabla E_{NN}(x_{NN}^*)=0$, and
\begin{equation}
	\|x_{NN}^* - x^*\|_2\leq \frac{4\varepsilon}{\mu}.
\end{equation}
\end{thm}

\begin{pro}
We define a function as 
\begin{equation}
	g(x)=x-(\nabla^2 E_{NN}(x^*))^{-1}\nabla E_{NN}(x).
\end{equation}
According to Corollary \ref{coro:4.5}, we have
\begin{align*}
	\|g(x^*) - x^*\|_2
	&\leq\|(\nabla^2 E_{NN}(x^*))^{-1}\|_2 \|\nabla E(x^*) + \nabla E_\delta(x^*)\|_2\\
	&\leq \frac{\varepsilon}{\mu - \varepsilon} \leq \frac{2\varepsilon}{\mu}
\end{align*}
and for any $x \in U(x^*, \delta)$,
\begin{align*}
	\|\nabla g(x)\|_2
	&= \|I-(\nabla^2 E_{NN}(x^*))^{-1}\nabla^2 E_{NN}(x)\|_2 \\
	&=\|(\nabla^2 E_{NN}(x^*))^{-1}(\nabla^2 E_{NN}(x^*)-\nabla^2 E_{NN}(x))\|_2 \\
	&\leq \|(\nabla^2 E_{NN}(x^*))^{-1}\|_2\left(\|\nabla^2 E(x^*)-\nabla^2 E(x)\|_2 +
	\|\nabla^2 E_{\delta}(x^*)-\nabla^2 E_{\delta}(x)\|_2\right) \\
	&\leq \frac{M+\varepsilon}{\mu-\varepsilon}\|x-x^*\|_2 \leq \frac{4M}{\mu}\|x-x^*\|_2.
\end{align*}
Then for $x \in U(x^*, D)$ and
\begin{equation}
	D=\frac{\mu-\sqrt{\mu^2-32M\varepsilon}}{8M}
	=\frac{4\varepsilon}{\mu+\sqrt{\mu^2-32M\varepsilon}}
	\leq \frac{4\varepsilon}{\mu}\leq \frac{\delta}{3},
\end{equation}
we have
\begin{align*}
	\|g(x) - x^*\|_2
	&\leq \|g(x^*) - x^*\|_2+\|g(x)-g(x^*)\|_2 \\
	&\leq \|g(x^*) - x^*\|_2 + \|\nabla g(x_{\xi})\|_2 \|x-x^*\|_2 \\
	&\leq \frac{2\varepsilon}{\mu} + \frac{4M}{\mu}D^2 = D
\end{align*}
and
\begin{equation*}
	\|\nabla g(x)\|_2 \leq \frac{4M}{\mu} D \leq \frac{16M\varepsilon}{\mu^2}
	\leq \frac{1}{2}.
\end{equation*}
So, $g(U(x^*, D)) \subset U(x^*, D)$ and $g(x)$ is a contraction mapping.
According to Banach fixed-point theorem, There exists an $x_{NN}^*\in U(x^*, D)$, such that
\begin{equation*}
	g(x_{NN}^*)=x_{NN}^*.
\end{equation*}
From the definition, it follows that
\begin{equation*}
	\nabla E_{NN}(x_{NN}^*)=0
\end{equation*}
and
\begin{equation*}
	\|x_{NN}^*-x^*\|_2\leq D \leq \frac{4\varepsilon}{\mu}.
\end{equation*}
\end{pro}

Now we can obtain the convergence analysis of the NN-HiSD method.
\begin{thm}
\label{thm:4.8}
Under Assumptions of Theorem \ref{thm:4.6}, if the initial point $x^{(0)}$ satisfies
\begin{equation*}
	\|x^{(0)}-x^*\|_2<min\{\frac{\delta}{3}, \frac{\hat{r}}{2}\}, \quad \hat{r}=\frac{\mu}{2M}.
\end{equation*}
and $r_0=\|x^{(0)}-x_{NN}^*\|_2, \ \beta_n=\frac{2}{L+\mu}$ for any $n\geq0$,
then $x^{(n)}$ converges to $x_{NN}^*$ as $n\to\infty$ with the estimate on the convergence rate
\begin{equation}
	r_n=\|x^{(n)}-x_{NN}^*\|_2\leq (1-\frac{2}{\tilde{\kappa}+3})^n\frac{\hat{r}r_0}{\hat{r}-r_0},
	\quad \tilde{\kappa}=\frac{L+\varepsilon}{\mu-\varepsilon}.
\end{equation}
\end{thm}

\begin{pro}

Firstly,
\begin{equation*}
	\|x^*-x_{NN}^*\|_2\leq\frac{4\varepsilon}{\mu}\leq min\{\frac{\delta}{3}, \frac{\hat{r}}{2}\},
	\ \hat{r}=\frac{\mu}{2M}.
\end{equation*}

Then for all $x,y\in U(x_{NN}^*,\frac{2\delta}{3})\subset U(x^*, \delta)$,
\begin{equation*}
	\|\nabla^2 E_{NN}(x) - \nabla^2 E_{NN}(y)\|_2\leq (M+\varepsilon)\|x-y\|_2\leq 2M\|x-y\|_2.
\end{equation*}

According to Corollary \ref{coro:4.5},
for any $x \in U(x_{NN}^*,\frac{2\delta}{3})$, eigenvalues $\{\lambda_i\}_{i=1}^d$ of $\nabla^2 E_{NN}(x)$
satisfy $\lambda_1 \leq \cdots \leq \lambda_k < 0 < \lambda_{k+1}\leq \cdots \leq \lambda_d$
and $|\lambda_i|\in [\mu-\varepsilon, L+\varepsilon]$ for $1\leq i \leq d$.

Besides, the initial point $x^{(0)}$ satisfies
\begin{equation*}
	r_0=\|x^{(0)}-x_{NN}^*\|_2<min\{\frac{2\delta}{3}, \hat{r}\},
	\ \hat{r}=\frac{\mu}{2M}.
\end{equation*}

According to the Lemma \ref{lem:4.3}, $\beta_n=\frac{2}{L+\mu}$ and
\begin{equation*}
	r_n=\|x^{(n)}-x_{NN}^*\|_2\leq (1-\frac{2}{\tilde{\kappa}+3})^n\frac{\hat{r}r_0}{\hat{r}-r_0},
	\quad \tilde{\kappa}=\frac{L+\varepsilon}{\mu-\varepsilon}.
\end{equation*}
\end{pro}

Theorem \ref{thm:4.8} shows that the convergence rate of NN-HiSD is \(1-\mathcal{O}\left(\frac{1}{\tilde{\kappa}}\right)\). This suggests that, upon the surrogate model being adeptly trained, the convergence velocity of the HiSD method employing the surrogate model predominantly remains unaltered.

Having established the convergence of the NN-HiSD method, it is also a logical progression to affirm the convergence of $\text{NN-HiSD}_\text{Hb}$ by employing \textbf{Theorem 4.6} in \cite{luo2023accelerated}.

\begin{coro}
We suppose Assumption \ref{assu:4.1}, \ref{assu:4.4}  hold and $\varepsilon$ is sufficiently small. We set $x^{(-1)}=x^{(0)}$ and 
\begin{equation*}
\sqrt{\beta_n} = \frac{2}{\sqrt{L+\varepsilon}+\sqrt{\mu-\varepsilon}}, \ \sqrt{\gamma}=1-\frac{3}{2(\sqrt{\tilde{\kappa}}+1)}, \  \tilde{\kappa}=\frac{L+\varepsilon}{\mu-\varepsilon}
\end{equation*}
in $\text{NN-HiSD}_\text{Hb}$. If the initial point $x^{(0)}$ satisfies that $\|x^{(0)}-x^*\|_2$ is sufficiently small, then $x^{(n)}$ converges to $x^*_{NN}$ as $n\to\infty$ with the estimate on the convergence rate
\begin{equation}
\|x^{(n)} - x_{NN}^*\|_2 \leq \tilde{K}\|x^{(0)} - x_{NN}^*\|_2(1 - \frac{1}{\sqrt{\tilde{\kappa}} + 1})^n,
\end{equation}
where $\tilde{K}$ is a positive constant with respect to $M, L, \mu, \delta$.
\end{coro}

\section{Numerical Experiments}
In this section, we conduct a series of experiments to test the NN-HiSD method with momentum acceleration. 
We use the $tanh$ activation function in all experiments, and the neural network of the surrogate model is trained using the Adam optimizer \cite{Kingma2014} with an initial learning rate of $lr=0.001$. 
Additionally, we employ a StepLR learning rate scheduler \cite{pytorch2023steplr} with a step size of 2000 and a decay factor of $0.7$ to further optimize the training process.

\subsection{2D and 3D Toy Models}
We employ the 2D and 3D toy models to evaluate the efficacy and discrepancies of the HiSD method with momentum acceleration using the exact potential function and its neural network-based surrogate model. 
A simple gradient system is given by
\begin{equation}\label{5.1}
  E(x)=\frac{1}{2}x^TMx-\alpha R(x).
\end{equation}

We first conduct a 2D case:
\begin{equation}
	M=\left(\begin{matrix}
	0.8 & -0.2 \\
	-0.2 & 0.5
	\end{matrix}\right), \quad R(x)=\sum_{i=1}^2\arctan(x_i-5), \quad \alpha=5.
\end{equation}

In the current experiment, we train the surrogate model using 5000 randomly selected data points from the domain $[-1,7] \times [-1,7]$. This neural network has three hidden layers, each equipped with 128 neurons. We implement training in batches of 500 over the course of 30000 epochs to ensure the neural network reaches optimal training levels. After finishing this phase, we move on to the next steps in the HiSD framework. 

For both the true potential and our surrogate model, we implement $\text{HiSD}_\text{Hb}$ with $\gamma=0.8$ and $\text{HiSD}_\text{NA}$ restarting every 20 steps. We search for index-1 saddle point from the initial point $(0.7, 0.7)$.  The time step $\beta_n$ is set $0.05$.

\begin{figure}[H]
\centering
\includegraphics[width=0.9\textwidth]{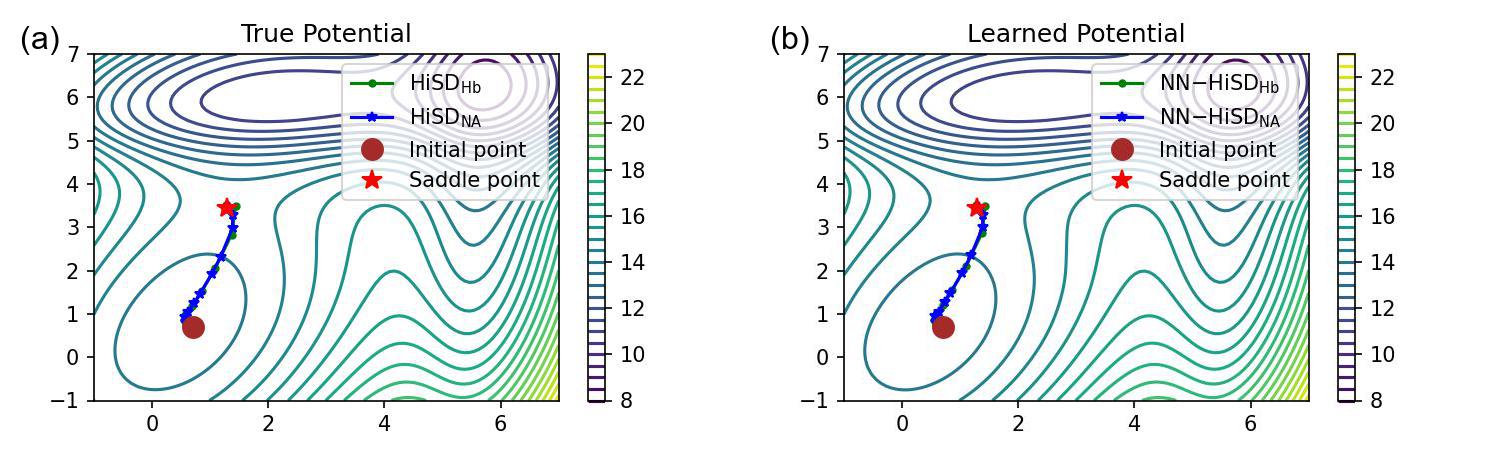}
\caption{Trajectory of true potential and learned potential. (a) true energy potential $E(x)$. (b) learned energy potential. The initial point is represented by a brown solid circle, and the saddle point is represented by a red solid pentagram. }
\label{fig2}
\end{figure}

The trajectory resulting from these settings is presented in Fig. \ref{fig2}, with progress plotted at intervals of every 10 steps. In our calculations, the true saddle point is $(1.2842, 3.4484)$, while the saddle point computed using the NN-HiSD method is $(1.2887, 3.4433)$, resulting in an error on the order of $10^{-3}$. From the results, we can conclude that the NN-HiSD with momentum acceleration performs equivalently to the HiSD method using exact potential function.

\begin{figure}[H]
\centering
\includegraphics[width=0.9\textwidth]{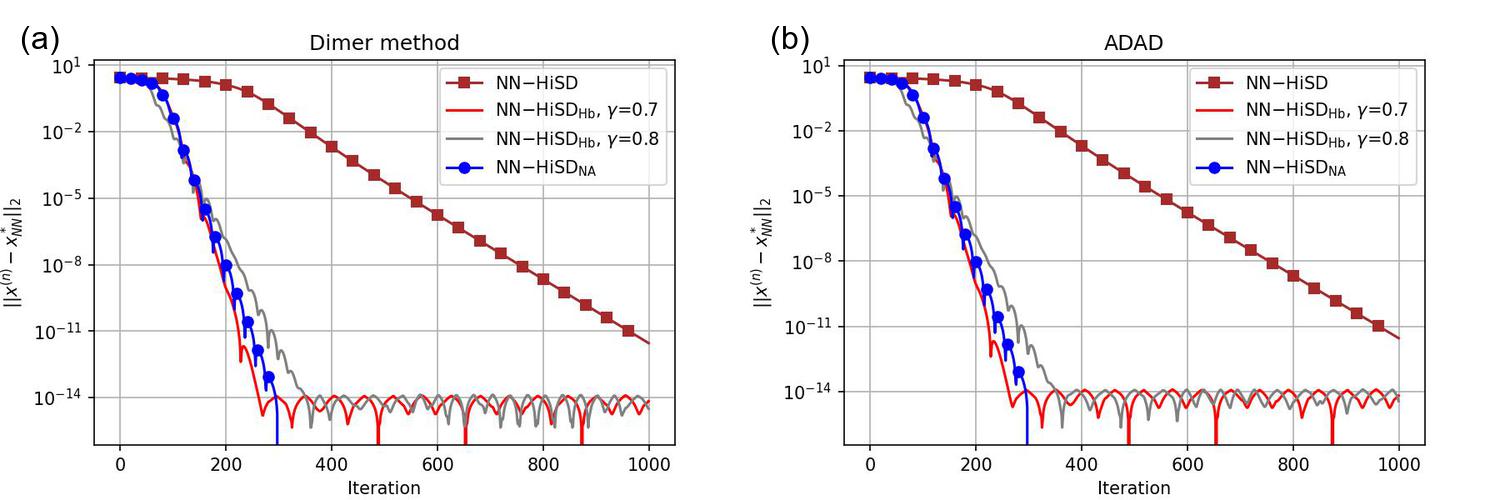}
\caption{$\|x^{(n)}-x_{NN}^*\|_2$ of the surrogate model-based HiSD method with respect to the iteration number. (a) dimer method to calculate $G(x)v_i$. (b) ADAD method to calculate $G(x)$.}
\label{fig3}
\end{figure}

We also compare the dimer method \cite{doi:10.1063/1.480097,doi:10.1137/19M1253356} and Automatic Differentiation (denoted as ADAD because we use automatic difference twice to calculate the corresponding Hessian matrix) in the computation of $G(x)v_i$ in \eqref{2.1b}. 
Convergence behaviors during the iteration of the HiSD are plotted in Fig. \ref{fig3}, demonstrating that the natural Automatic Differentiation technique of the neural network can fully replace the dimer method to achieve satisfactory results.

In practical scenarios, the training data we acquire typically includes some degree of noise, thus necessitating an assessment of the surrogate model's resistance to noise. To this end, we utilize a set of 5000 constant data points within the region $[-1,7] \times [-1,7]$ as the inputs for the surrogate model, introducing Gaussian noise with a standard deviation of 0.1 to a subset of these points. The architecture of the surrogate model involves a neural network with three hidden layers, each containing 128 neurons. We train the model in batches of 500 across 30000 epochs.

In this designated region, we identify two index-1 saddle points: Saddle1 at coordinates $(1.2842, 3.4484)$ and Saddle2 at $(3.5689, 6.0735)$. Utilizing initial points at $(1.3, 3.5)$ and $(3.5, 6.0)$, we employ the $\text{NN-HiSD}_\text{NA}$ method, which includes a procedure of restarting every 500 steps, with a setting of $\beta = 10^{-4}$. This approach successfully locates the index-1 saddle point within 3000 steps. 
We define the error as 
$$
\text{Error} = \|u_e - u_s\|,
$$
where $u_e$ represents the exact saddle point location and $u_s$ denotes the saddle point calculated by the surrogate model. The findings are presented in Table \ref{table1}.

The term ``Noise Ratio" denotes the fraction of the training data to which Gaussian noise has been added. According to the data presented in this table, it becomes evident that our methodology demonstrates robustness in calculating saddle points.
\begin{table}[H]
\caption{Errors with different ratios of the training data that has Gaussian noise added}
\label{table1}
\centering
\begin{tabularx}{0.7\textwidth}{>{\centering\arraybackslash}X>{\centering\arraybackslash}X>{\centering\arraybackslash}X}
\toprule
\textbf{Noise Ratio} & \textbf{Error of Saddle1} & \textbf{Error of Saddle2} \\
\midrule
0$\%$ & 0.007753 & 0.001658 \\
$10\%$ & 0.050641 & 0.036997 \\
$20\%$ & 0.064979 & 0.033291 \\
$30\%$ & 0.103624 & 0.071350 \\
$40\%$ & 0.096967 & 0.089498 \\
$50\%$ & 0.100412 & 0.112222 \\
$60\%$ & 0.182844 & 0.183391 \\
$70\%$ & 0.199748 & 0.125200 \\
$80\%$ & 0.207824 & 0.186855 \\
$90\%$ & 0.438199 & 0.452666 \\
$100\%$ & 0.702949 & 0.507552 \\
\bottomrule
\end{tabularx}
\end{table}

Next is a 3D parametric case that is given by
\begin{equation}
E(x)=\frac{1}{2}x^TMx-\alpha R(x), \quad
M=\left(\begin{matrix}
	0.8 & -0.2 & -0.5 \\
	-0.2 & 0.5 & 0.2 \\
	-0.5 & 0.2 & 1
\end{matrix}\right), \quad R(x)=\sum_{i=1}^3\arctan(x_i-5).
\end{equation}
Here the parameter $\alpha$ is a variable, and it is considered as one of inputs of the neural network, which consists of three hidden layers, each with 128 neurons. 
Specifically, for the parameter \(\alpha\), 10 different values are chosen, evenly spaced within the interval \([3, 7]\). For each specific \(\alpha\), we randomly select 4000 data points within the region \([-1,7] \times [-1,7] \times [-1,7]\) as training data. Therefore, a total of 40000 training data points are used to fit the surrogate model. 
The model is trained with a batch size of 1000 for 30000 epochs. 

\begin{figure}[hbtp]
\centering
\includegraphics[width=0.9\textwidth]{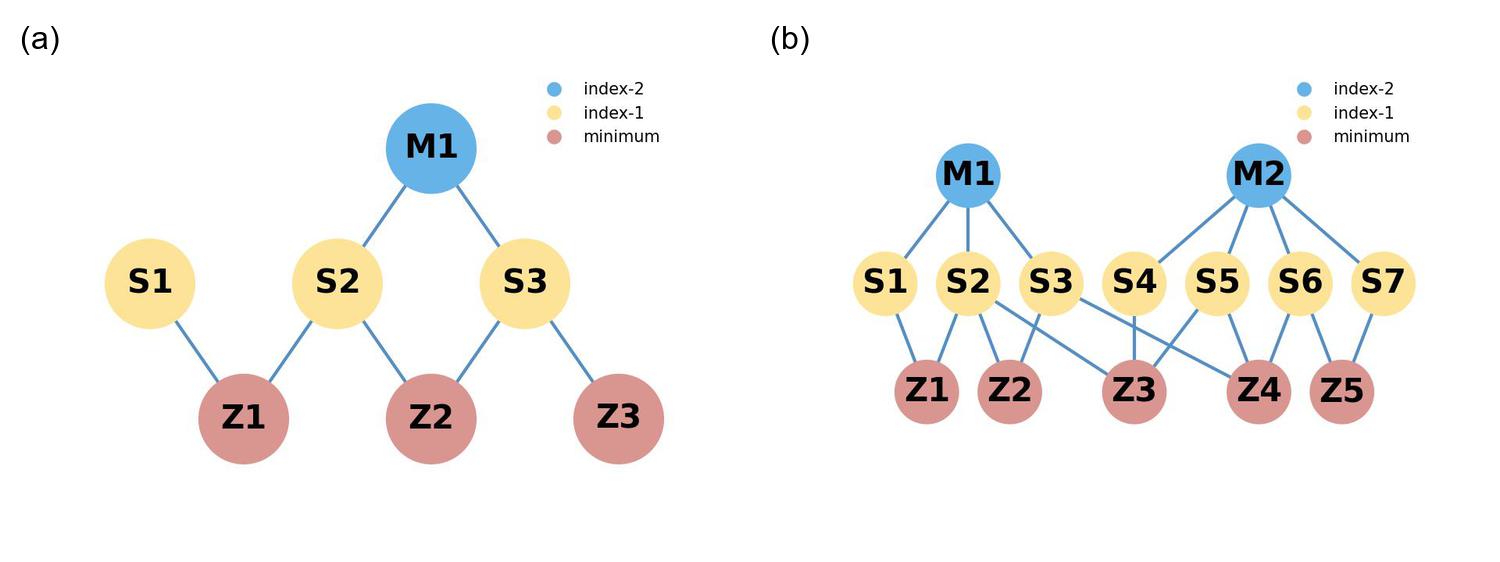}
\caption{Solution landscape for both true model and surrogate model. (a) $\alpha=6$ \ (b) $\alpha=3$}
\label{fig4}
\end{figure}

In order to validate the performance of the surrogate model, we construct the solution landscape of both true model and surrogate parametric model for certain $\alpha$ in Fig. \ref{fig4}, and present values of corresponding saddle points in Table \ref{table2}. `M' denotes an index-2 saddle, `S' denotes an index-1 saddle, and `Z' denotes a minimum. It turns out that the true model and the surrogate parametric model share the same solution landscape. This indicates that we can use the NN-HiSD to construct the solution landscape easily and accurately.

{\tiny
\begin{table}[H]
\begin{minipage}[b]{0.5\textwidth}
\setlength{\abovecaptionskip}{0.33cm}
\caption{Coordinates of saddle points for different $\alpha$ values, (a)  $\alpha=6$ and (b) $\alpha=3$. If $/$ appears, then the former is the true saddle point and the latter is the calculated saddle point. If it does not appear, this indicates that the calculated saddle point coincides with the true saddle point.}
\label{table2}
\centering
\scalebox{0.9}{ 
\begin{tabular}{cccc}
\hline
\textbf{Saddle} & $x_1$ & $x_2$ & $x_3$ \\
\hline
M1 & 2.89/2.88 & 2.94 & 1.25 \\
S1 & 6.61 & 3.26 & 5.92 \\
S2 & 3.10 & 1.35 & 1.82 \\
S3 & 1.82/1.80 & 3.10 & 0.58/0.57 \\
Z1 & 6.43 & 0.80/0.81 & 6.01 \\
Z2 & 1.10 & 0.80/0.81 & 0.69 \\
Z3 & 6.85 & 6.06 & 5.81/5.82 \\
\hline
\end{tabular}}
\caption*{(a)}
\end{minipage}%
\begin{minipage}[b]{0.5\textwidth}
\centering
\scalebox{0.9}{
\begin{tabular}{cccc}
\hline
\textbf{Saddle} & $x_1$ & $x_2$ & $x_3$ \\
\hline
M1 & 4.17/4.18 & 3.97/3.98 & 1.52/1.53 \\
M2 & 5.91 & 4.19 & 4.48 \\
S1 & 1.22/1.23 & 4.20 & -0.12/-0.11 \\
S2 & 4.34 & 1.21 & 2.28 \\
S3 & 4.05 & 5.60 & 1.09/1.08 \\
S4 & 5.48 & 3.86 & 2.34 \\
S5 & 5.55 & 1.05 & 3.83 \\
S6 & 6.05/6.06 & 4.26 & 5.13 \\
S7 & 6.07/6.06 & 5.47/5.46 & 4.72/4.71 \\
Z1 & 0.42/0.43 & 0.32/0.31 & 0.28 \\
Z2 & 1.51 & 5.50 & -0.24/-0.25 \\
Z3 & 5.43 & 1.36 & 3.09/3.08 \\
Z4 & 5.53 & 5.65 & 1.92 \\
Z5 & 5.79 & 0.43 & 5.40 \\
Z6 & 6.12/6.13 & 5.45 & 4.97/4.98 \\
\hline
\end{tabular}}
\caption*{(b)}
\end{minipage}
\end{table}
}

\subsection{M{\"{u}}ller-Brown Potential}
In the second experiment, we turn to the well known M{\"{u}}ller-Brown (MB) potential, a benchmark to test the performance of saddle point searching algorithms.
The MB potential is given by \cite{doi:10.1063/1.5012271}
\begin{equation}
E_{M B}(x, y)=\sum_{i=1}^4 A_i \exp \left[a_i\left(x-\bar{x}_i\right)^2+b_i\left(x-\bar{x}_i\right)\left(y-\bar{y}_i\right)+c_i\left(y-\bar{y}_i\right)^2\right],
\end{equation}
where $
A = [-200, -100, -170, 15],~
a = [-1, -1, -6.5, 0.7],~
b = [0, 0, 11, 0.6],~\\
c = [-10, -10, -6.5, 0.7],~
\bar{x} = [1, 0, -0.5, -1],~
\bar{y} = [0, 0.5, 1.5, 1].
$
We also use modified M{\"{u}}ller-Brown (MMB) potential
\begin{equation}
E_{M M B}(x, y)=E_{M B}(x, y)+A_5 \sin (x y) \exp \left[a_5\left(x-\bar{x}_5\right)^2+c_5\left(y-\bar{y}_5\right)^2\right] .
\end{equation}
where $A_5=500,~ a_5=-0.1,~ c_5 =  -0.1,~
\bar{x}_5=-0.5582,~ \bar{y}_5 = 1.4417.$

The surrogate model is trained by a neural network consists of three hidden layers with each 128 neurons. We use 5000 random data points in the region $[-1.5,0.25]\times[0.0,1.95]$ and their potential values as training data. The surrogate model is trained with a batch size of 250 for 30000 epochs.
Once the neural network is well-trained, we implement NN-HiSD, $\text{NN-HiSD}_\text{Hb}$ with $\gamma=0.8$ and $\text{NN-HiSD}_\text{NA}$ for index-1 saddle point from the initial point $(0.15, 0.25)$, setting $\beta=3\times 10^{-4}$. For comparison, we also applied the HiSD method with the same parameters to the true potential energy function.    

\begin{figure}[htbp]
\centering
\includegraphics[width=0.9\textwidth]{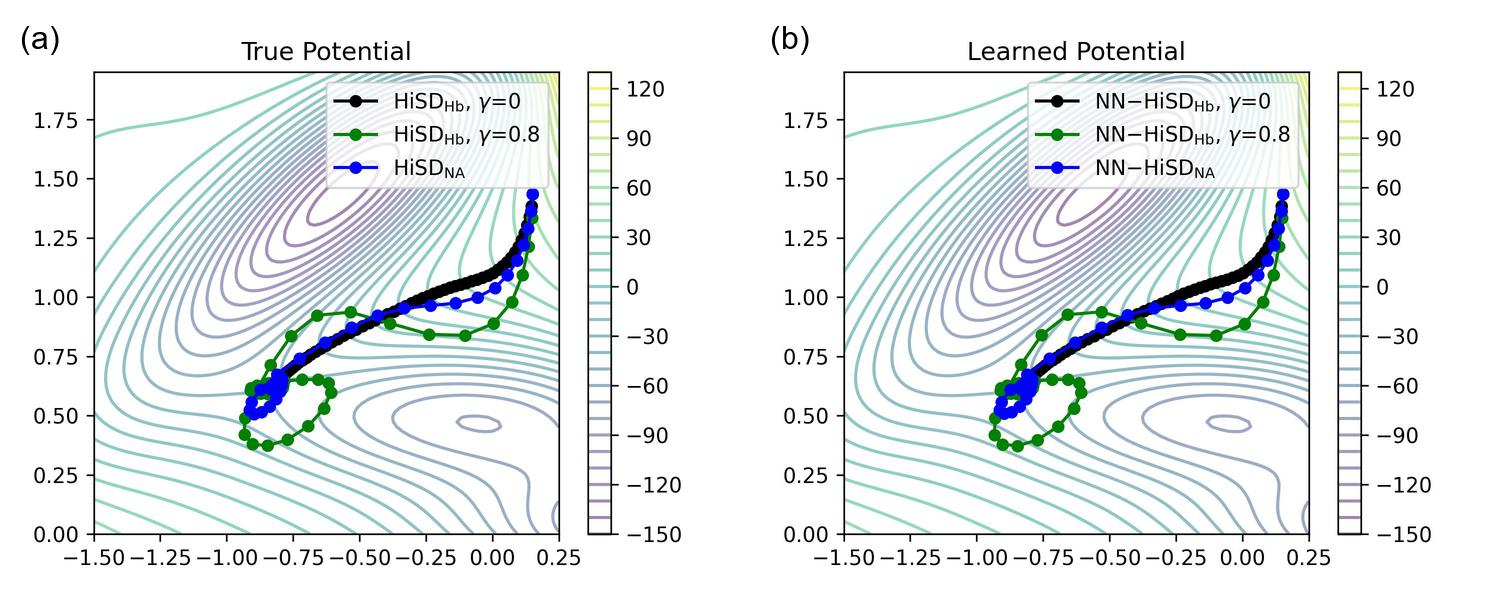}
\caption{Trajectory of true potential and learned potential for the MB function. (a) True energy potential $E(x)$. (b) Learned energy potential $E_{NN}(x)$.}
\label{fig5}
\end{figure}

Fig. \ref{fig5} shows the results, in which the progress is plotted at every iteration. In our calculations, the index-1 saddle point of MB potential is $(-0.8220, 0.6243)$, while the index-1 saddle point computed using the NN-HiSD method is $(-0.8216, 0.6243)$, with an error on the order of $10^{-4}$. We can see that the surrogate models can closely mimic the exact scenarios.

\begin{figure}
\centering
\includegraphics[width=0.9\textwidth]{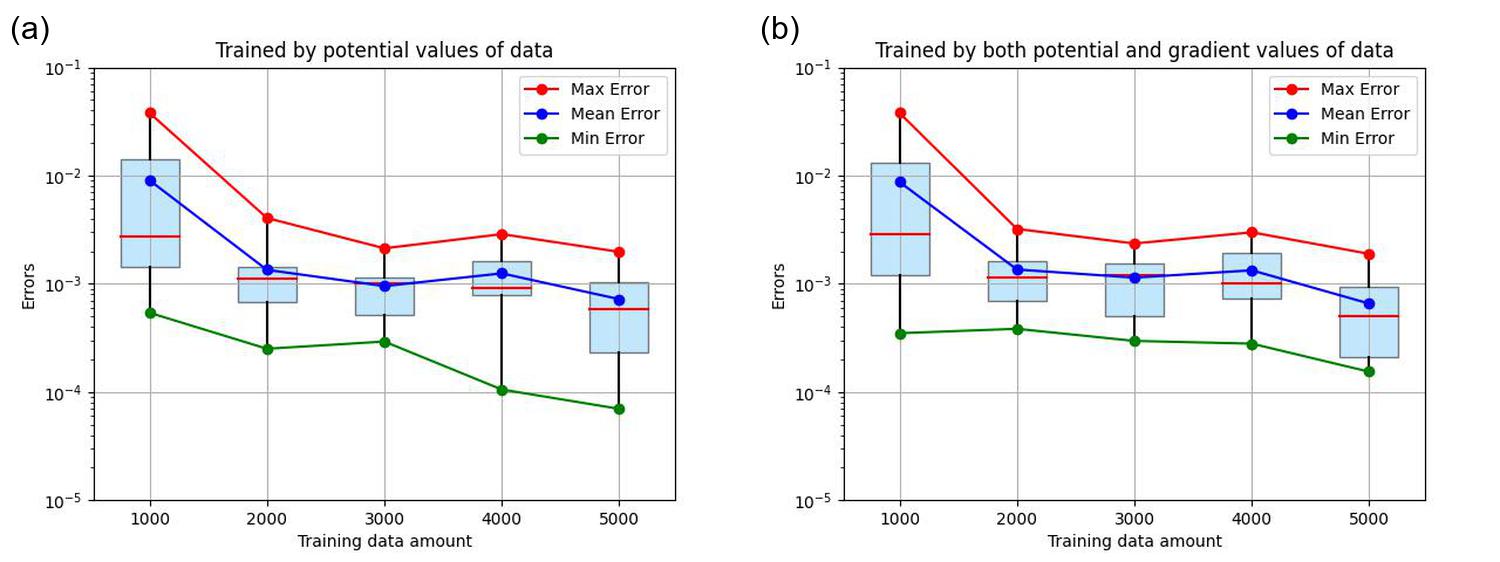}
\caption{Boxplot illustrating saddle point prediction errors across various amounts of training data, 
tested across $10$ different training data distributions for every amount. 
The upper and lower edges of each box indicate the upper and lower quartiles, respectively, 
while the central red line denotes the median. 
The maximum, average, and minimum errors are represented by three separate lines.
(a) Based solely on energy potential. (b) based on both potential and gradient information for 15\% of the points.}
\label{fig6}
\end{figure}

We also investigate how the amount of training data affects the surrogate model's performance in predicting saddle points. The surrogate model is trained as before, but with different training data sizes.
The results are shown in Fig. \ref{fig6}(a). Clearly, more data points enhance model robustness and accuracy.

Besides, We investigate the impact of incorporating gradients during training on the surrogate model in \ref{fig6}(b).
After training for 30000 epochs, we further incorporate the gradient error at 15\% of the points into the training loss 
and continue training for an additional 5000 epochs. We observe that incorporating gradients can enhance the model's robustness, indicating that including gradients during training is indeed an effective approach.

\begin{figure}[hbtp]
\centering
\includegraphics[width=0.9\textwidth]{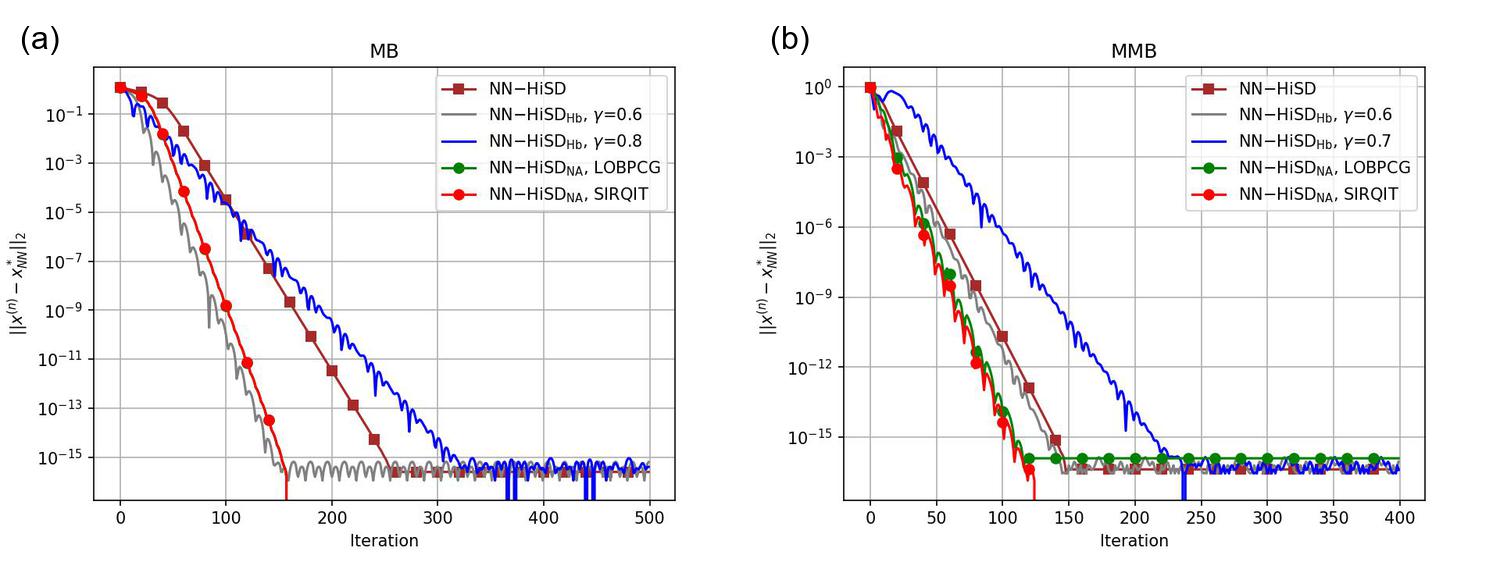}
\caption{Plots of $\|x^{(n)}-x_{NN}^*\|$ for MB and MMB potential in various NN-HiSD procedure. (a) MB potential. (b) MMB potential.}
\label{fig7}
\end{figure}

In addition, we compare the acceleration of $\text{NN-HiSD}_\text{Hb}$ and $\text{NN-HiSD}_\text{NA}$ for MB and MMB potentials in Fig. \ref{fig7}. We also evaluate how the eigenvalue \( v_i \) computations affect convergence, comparing SIRQIT and LOBPCG.

Both surrogate models are trained using a neural network with three hidden layers of $128$ neurons each, on 10000 data points, a batch size of 1000, and over 30000 epochs. After the training stage, we search for index-1 saddle point with $\beta = 3 \times 10^{-4}$ for all methods. For MB, the region is $[-1.5,0.25]\times[0.0,1.95]$, the initial point is $(0.15, 1.5)$, and $\text{NN-HiSD}_\text{NA}$ restarts every 5 steps. For MMB, the region is $[-2.8, 0.89]\times[0.0, 2.2]$, the initial point is $(0.15, 1.4)$, and $\text{NN-HiSD}_\text{NA}$ restarts every 15 steps.

It turns out that the convergence rates of $\text{NN-HiSD}_\text{Hb}$ and $\text{NN-HiSD}_\text{NA}$ are similar, but $\gamma$ does not need to be adjusted for $\text{NN-HiSD}_\text{NA}$. The convergence rate of LOBPCG and SIRQIT are similar; however, SIRQIT proves to be faster in computation.

\subsection{Rosenbrock-type Function}
In the third experiment, we analyze a well-known test problem in optimization: the 
$d$-dimensional Rosenbrock-type function, defined as follows:
\begin{equation}
R(x)=\sum_{i=1}^{d-1}\left[100\left(x_{i+1}-x_i^2\right)^2+\left(1-x_i\right)^2\right].
\end{equation}

There are lots of saddle points of $R(x)$ around $x^*=(1, \cdots, 1)$, so we use its modified form \cite{doi:10.1137/22M1487965}
\begin{equation}
R_m(x)=R(x)+\sum_{i=1}^d s_i \arctan ^2\left(x_i-x_i^*\right).
\end{equation}
Here the parameters are set as $d = 7,~ s = [-50000, -50000, -50000, -50000, -50000, 1, 1]$. It can be proved that $x^*$ is an index-$3$ saddle point of the modified Rosenbrock function.

The surrogate models are trained on different numbers of random data points within the region $[0.8, 1.2]^7$, utilizing a neural network with three hidden layers of 256 neurons each, and trained with a batch size of 100 for 30000 epochs. We employ the $\text{NN-HiSD}_\text{NA}$ method, restarting every 40 steps, with the initial point $(0.9, \ldots, 0.9)$ and $\beta = 10^{-4}$. 

Table \ref{table3} displays index-3 saddle points of surrogate models for various numbers of sampling points, in which  ``Data numbers'' means the number of the training data and ``Saddle point'' means the calculated saddle point. The $\text{NN-HiSD}_\text{NA}$ method converges within $2000$ iterations for all experiments. The results suggest that increasing the number of data points can enhance accuracy.

\begin{table}[H]
\centering{}
\caption{Index-3 saddle points computed by $\text{NN-HiSD}_\text{NA}$ where the surrogate model is trained on different data points.}
\resizebox{0.9\textwidth}{!}{
\begin{tabular}{cc}
\hline
\textbf{Data numbers} & \textbf{Saddle point} \\
\hline
10000 & $(1.0007, 1.0006, 0.9998, 0.9998, 1.0003, 1.0006, 1.0016)$ \\
20000 & $(1.0001, 0.9999, 0.9999, 1.0000, 1.0002, 1.0000, 0.9998)$ \\
30000 & $(1.0004, 0.9998, 0.9998, 1.0000, 1.0004, 1.0000, 0.9997)$\\
40000 & $(1.0000, 0.9999, 0.9999, 0.9999, 1.0000, 0.9999, 1.0000)$ \\
50000 & $(1.0000, 0.9998, 0.9999, 1.0000, 1.0001, 0.9999, 1.0001)$ \\
\hline
\end{tabular}}
\label{table3}
\end{table}

\begin{figure}[htbp]
\centering
\includegraphics[width=0.6\textwidth]{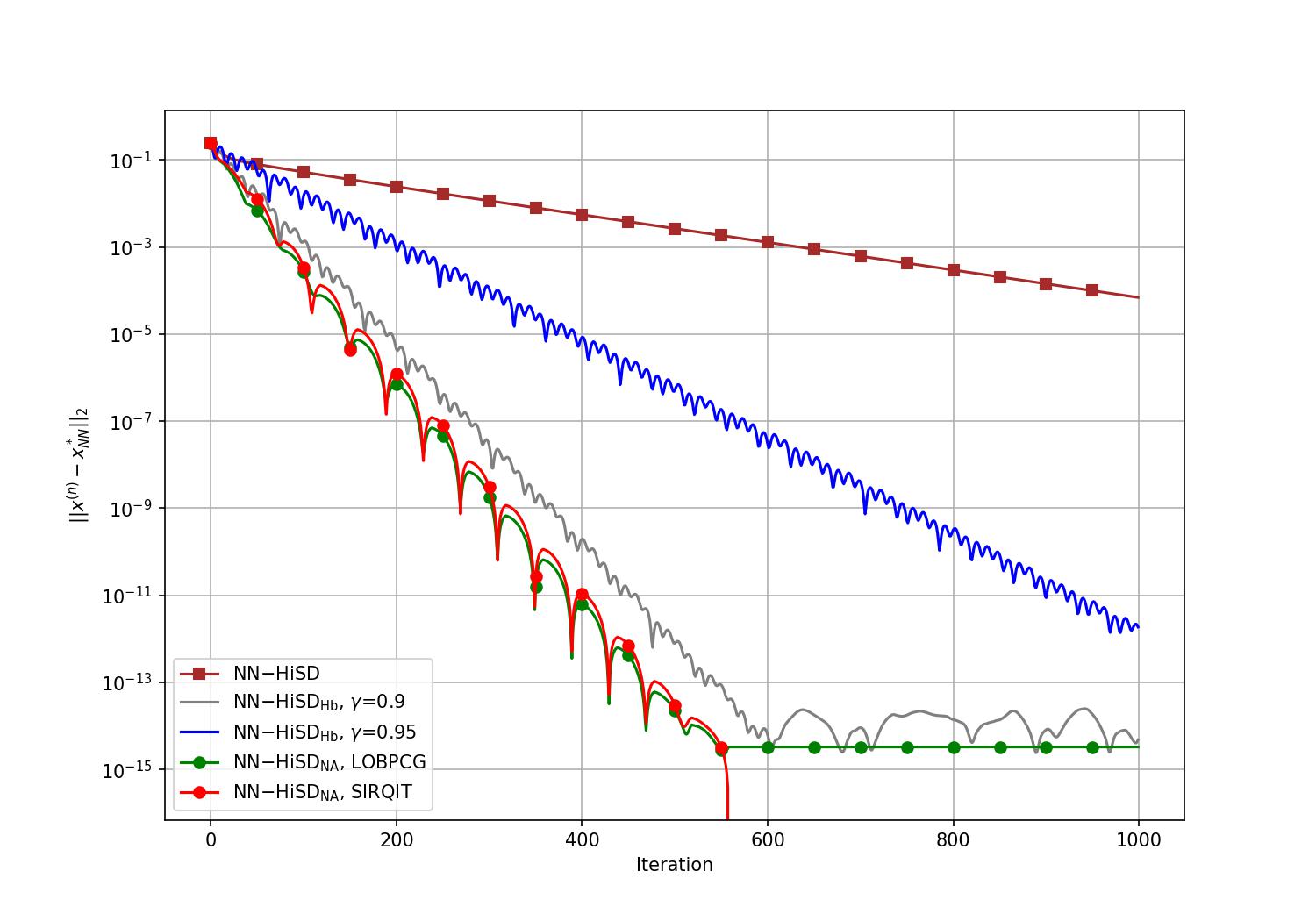}
\caption{Plots of $\|x^{(n)}-x_{NN}^*\|$ in the first 1000 iterations of various NN-HiSD methods.}
\label{fig8}
\end{figure}

Similar to the previous experiment in Fig. \ref{fig7}, we compare the acceleration of various NN-HiSD methods and $v_i$ computations for the case of 50000 points in Table \ref{table3}. For all methods, the initial point is $(0.9, \cdots, 0.9)$ and $\beta=10^{-4}$. Results are displayed in Fig. \ref{fig8}.

\subsection{Alanine Dipeptide Model}
\begin{figure}[htbp]
\centering
\includegraphics[width=0.55\textwidth]{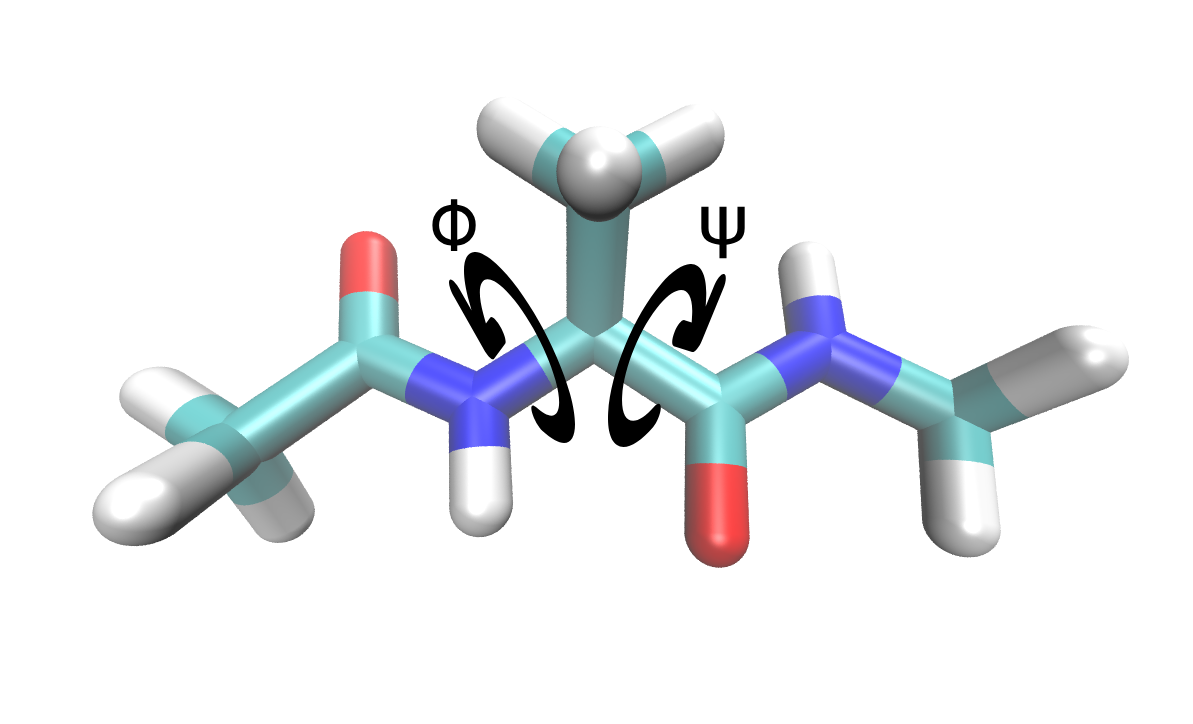}
\caption{Diagram of the alanine dipeptide (CH3–CONH–CHCH3–CONH–CH3)}
\label{alanine}
\end{figure}

In this example, we apply the NN-HiSD method to the $\Phi$-$\Psi$ dihedrals of alanine dipeptide, a 22-dimensional Molecular Dynamic model (Fig. \ref{alanine})\cite{doi:10.1007/s10915-022-02040-1,doi:10.1063/1.2013256}. 


We study the isomerization process of the alanine dipeptide in vacuum at $T = 300K$.
The free energy associated with $(\phi(x),\psi(x))$ is the function depending on
$v = (v_1, v_2)$, defined as
\begin{equation}
F(\boldsymbol{v})=-k_B T \ln \left(A^{-1} \int_{\mathbb{R}^d} e^{-\frac{V(\boldsymbol{x})}{k_B T}} 
\times \delta\left(v_1-\phi(\boldsymbol{x})\right) \times \delta\left(v_2-\psi(\boldsymbol{x})\right) d \boldsymbol{x}\right),
\end{equation}
where $A = \int_{\mathbb{R}^d} e^{-\frac{V(\boldsymbol{x})}{k_B T}} d\boldsymbol{x}$, 
$T$ is the temperature, $k_B$ is the Boltzmann constant, $\delta(\cdot)$ denotes the Dirac-delta function, 
and $V(\boldsymbol{x})$ represents the potential energy function of all atoms' positions $\boldsymbol{x} \in \mathbb{R}^d$.

It is very difficult to obtain the free energy for certain values of $\phi$ and $\psi$. Therefore, we use molecular dynamics (MD) simulations to obtain the free energy values with noise. We employ the NAMD \cite{Phillips2005} package to simulate Langevin dynamics, ultimately obtaining the free energy $F$ at $v=(\phi, \psi)$ 
within the set $\{(\phi, \psi) | -180\leq \phi, \psi<180, \phi, \psi \in \mathbb{Z}\}$. Metadynamics is utilized with a time step size of $0.5$ and $2\times10^7$ steps. 

The model is refined on a grid comprising $360 \times 360$ points utilizing a neural network that features five hidden layers, each layered with 128 neurons. This training session spans over 20000 epochs with batches of 1000. Due to the complexities associated with managing large-scale datasets, we undertake a preprocessing step to normalize the data, setting the mean to $0$ and standard deviation to $1$ prior to its utilization as input for the neural network. When constructing the solution landscape, the same transformation should be applied to both the input and output of the NN-HiSD method.

\begin{figure}[htbp]
\centering
\includegraphics[width=0.9\textwidth]{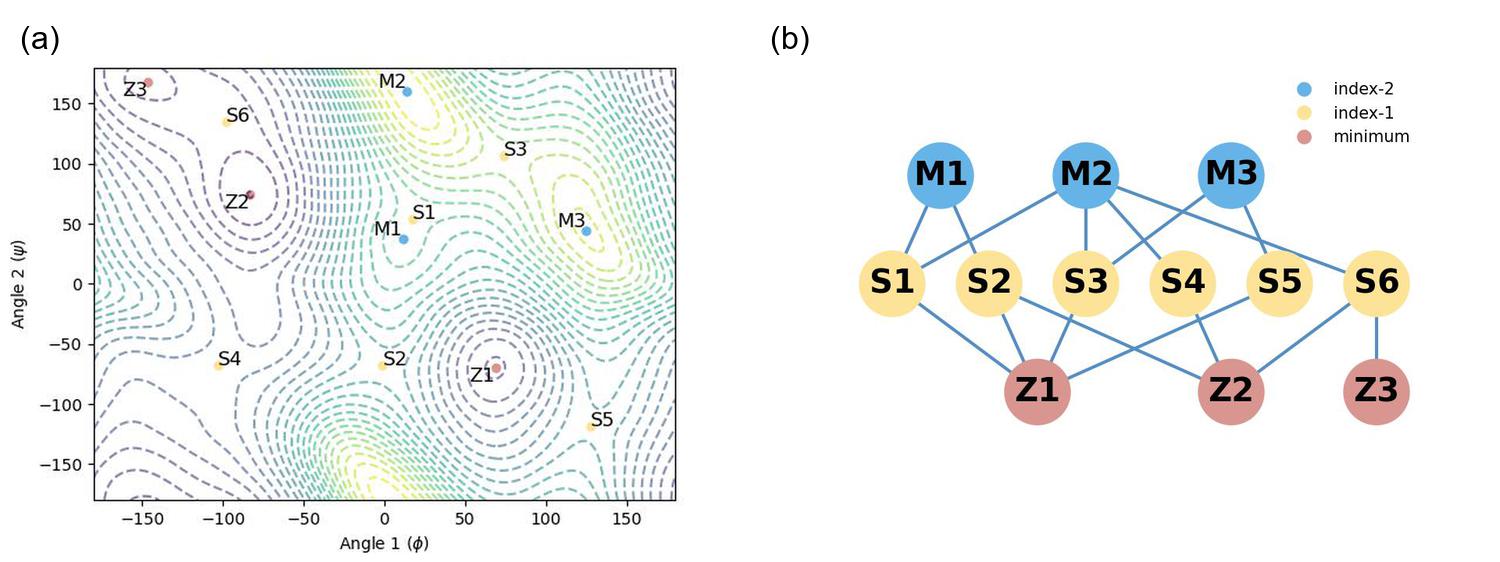}
\caption{(a) Contour and saddle points of the surrogate model. (b) Solution landscape of alanine dipeptide model.}
\label{fig9}
\end{figure}

The contours and critical points of the learned free energy landscape are illustrated in Fig. \ref{fig9}a. In this depiction, `M' denotes an index-2 saddle, `S' signifies an index-1 saddle, and `Z' denotes a minimum. Fig. \ref{fig9}b displays the solution landscape calculated by our surrogate model. The coordinates of the saddle points are presented in Table \ref{table4}. 

In each iteration of the NN-HiSD process, we utilize values from the surrogate model, which eliminates the need to reference the MD results and significantly reduces computation time. We calculate all possible critical points of the alanine dipeptide model, including the stable configurations $C_7^{ex}$, $C_7^{eq}$ and \(C_5\) \cite{Mironov2019nanma} (denoted as Z1, Z2 and Z3 in Fig. \ref{fig9}). The constructed solution landscape provides a more intuitive and comprehensive way to study the transition pathways of the alanine dipeptide model.
\begin{table}[H]
\caption{Coordinates of all saddle points calculated by the surrogate model of alanine dipeptide.}
\centering{}
\resizebox{\textwidth}{!}{
\begin{tabular}{ccccccccccccc}
\hline
\textbf{Saddle Points} & M1 & M2 & M3 & S1 & S2 & S3 & S4 & S5 & S6 & Z1 & Z2 & Z3 \\
\hline
$\phi$ & 11.3 & 14.0 & 125.2 & 17.6 & -1.1 & 73.8 & -103.2 & 127.4 & -98.1 & 68.7 & -83.2 & -146.4 \\
\hline
$\psi$ & 17.7 & 160.4 & 44.5 & 53.5 & -68.2 & 106.2 & -67.7 & -118.7 & 135.0 & -69.7 & 74.0 & 168.4 \\
\hline
\end{tabular}}
\label{table4}
\end{table}

\subsection{Global Pseudo-energy Landscape of Bacterial Ribosomal Assembly Intermediates}
\begin{figure}[htbp]
\centering
\includegraphics[width=0.9\textwidth]{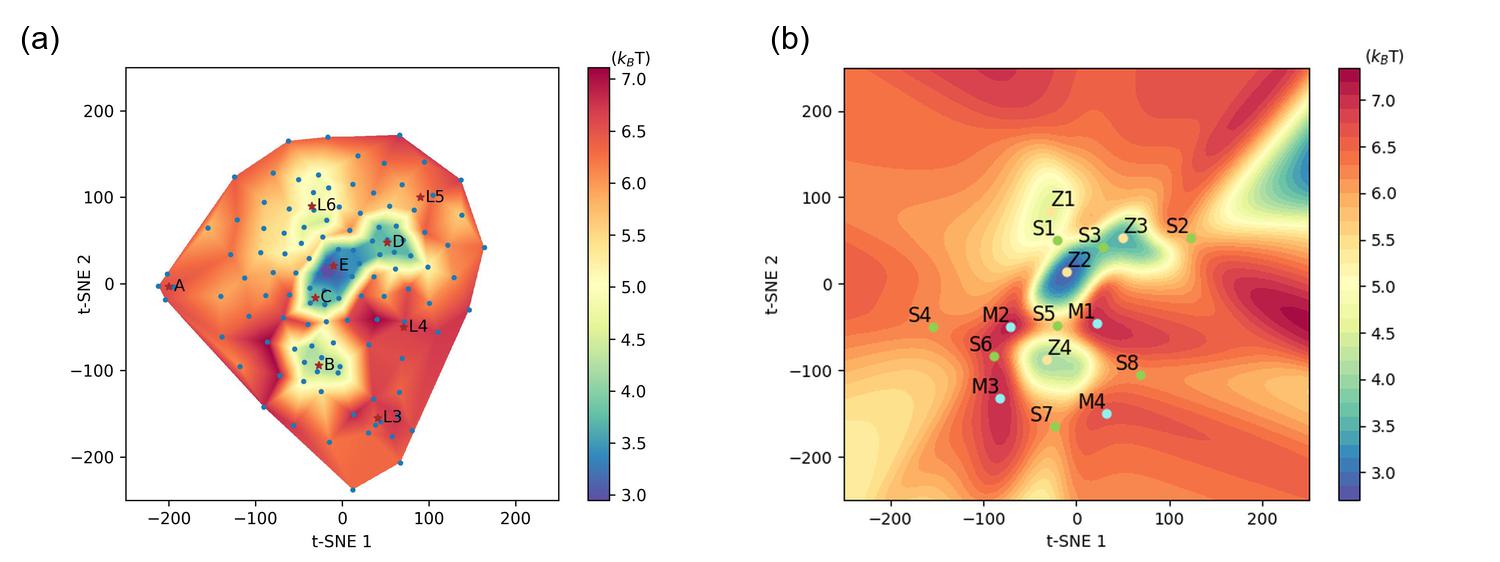}
\caption{(a) Global pseudo-energy landscape of bacterial ribosomal assembly intermediates. Clusters A-E and L3-L6 correspond to several important states.
(b) Contour and saddle nodes of the surrogate model approximating the energy landscape. }
\label{fig10}
\end{figure}

In the last experiment, we use one of the test cases of AlphaCryo4D \cite{ijms23168872} to analyse the overall pseudo-energy landscape of the ribosomal assembly using 119 density maps (Fig. \ref{fig10}a). A highly heterogeneous dataset of the \textit{Escherichia coli} 50S large ribosomal subunit (LSU) (EMPIAR-10076) \cite{doi:10.1016/j.cell.2016.11.020} is analyzed. This dataset (131,899 particles) is known to be both compositionally heterogeneous and conformationally dynamic, as the 50S LSU undergoes bL17-depleted intermediate assembly. Each saddle point in the landscape corresponds to a major conformational state. Therefore, constructing the solution landscape is necessary and valuable.

The surrogate model is trained on $119$ data points using a neural network with three hidden layer of 256 neurons each and a batch size of $5$ for 5000 epochs. With only 119 data points, overfitting is a significant concern during training. Hence, we introduce a regularization term with a regularization coefficient of $4\times10^{-3}$.  To enhance the model's performance, we also preprocess the data to have a mean of 0 and a standard deviation of 1. Since the input to the surrogate model has been preprocessed, the same transformation should be applied to the input and output of the NN-HiSD method. 

\begin{figure}[htbp]
\centering
\includegraphics[width=0.6\textwidth]{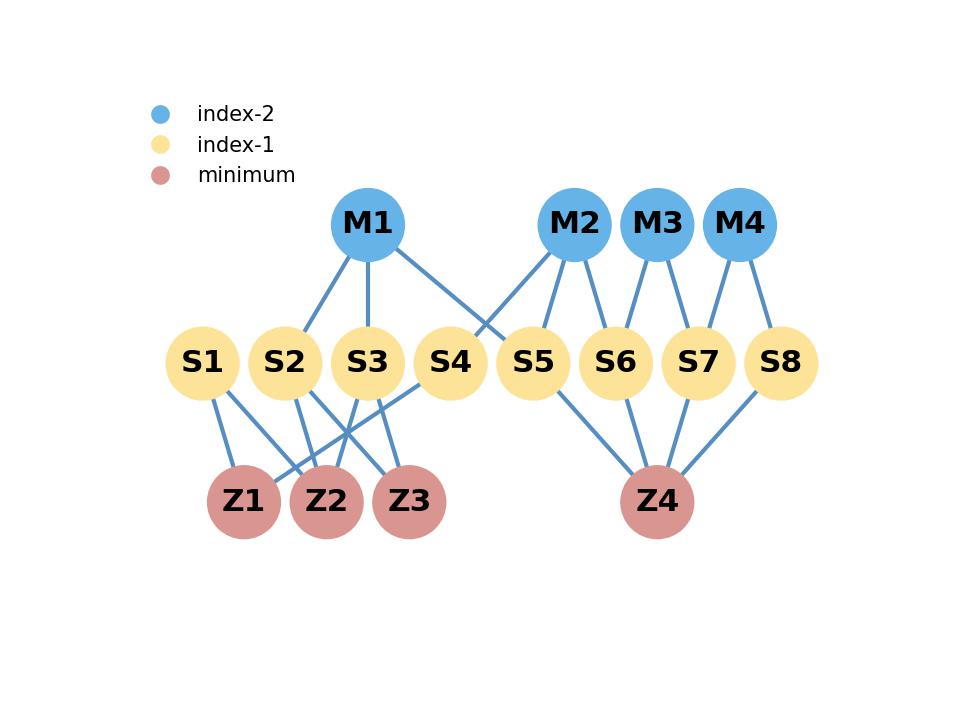}
\caption{Solution landscape calculated by the surrogate model.}
\label{fig11}
\end{figure}

The contour and saddle points of learned energy landscape is shown in Fig. \ref{fig10}b. 
`M' denotes an index-2 saddle, `S' signifies an index-1 saddle, and `Z' denotes a minimum. The solution landscape calculated by our surrogate model is shown in Fig. \ref{fig11}.  Coordinates of corresponding saddle points are shown in Table \ref{table5}.

We demonstrate that there exist some saddle points near clusters A-E and L3-L6. For example, the minimas Z1, Z2, Z3 and Z4 are located adjacent to clusters L6, E, D and B, respectively. This indicates that clusters A-E and L3-L6 are indeed significant states, consistent with the experimental findings in \cite{ijms23168872}.  Furthermore, the additional transition states like S1 and S3 have not been observed in the experiments, which can be considered as new potential conformational states. This observation also corroborates the feasibility and accuracy of the surrogate model.
\begin{table}[H]
\caption{Coordinates of all saddle points calculated by the surrogate model of the bacterial ribosomal assembly intermediates. }
\centering{}
\resizebox{\textwidth}{!}{
\begin{tabular}{ccccccccccccccccc}
\hline
\textbf{Saddle Point} & M1 & M2 & M3 & M4 & S1 & S2 & S3 & S4 & S5 & S6 & S7 & S8 & Z1 & Z2 & Z3 & Z4\\
\hline
\textbf{t-SNE 1} & 22 & -71 & -82 & 32 & -21 & 28 & 123 & -154 & -21 & -89 & -23 & 69 & -28 & -10 & 50 & -32\\
\hline
\textbf{t-SNE 2} & -46 & -49 & -132 & -149 & 51 & 43 & 53 & -50 & -48 & -83 & -165 & -105 & 85 & 14 & 53 & -88\\
\hline
\end{tabular}}
\label{table5}
\end{table}


\section{Conclusions}
In this paper, we introduce the NN-HiSD method, which utilizes a neural network surrogate model to approximate the energy function and employs the HiSD on this surrogate model to compute saddle points and solution landscapes. We further incorporate the momentum accelerated techniques, including Nesterov's acceleration and heavy-ball method, to improve the computational efficiency of the NN-HiSD method. The comprehensive proof of convergence established for the surrogate model framework substantiates the theoretical robustness and dependability of our method. We illustrate the reliability of the NN-HiSD method through various numerical experiments, including data-driven models like the alanine dipeptide and bacterial ribosomal assembly intermediates. Numerical results demonstrate that the proposed approach opens new avenues for exploring a wide range of complex systems. 

The NN-HiSD method proposed in this paper is primarily used for computing saddle points in gradient systems. For the non-gradient systems, we can extend the NN-HiSD method by directly fitting a surrogate model to the force information $F(x)$. Future work will focus on optimizing the surrogate model by using other neural network architectures and exploring more applications in complex systems.

\section*{Code and Data Availability}
Data and code are available from the corresponding author upon reasonable request
for research purposes.
\section*{Acknowledgements}
We thank Prof. Youdong Mao for sharing the data of the bacterial ribosomal assembly intermediates. We also thank Dr. Yue Luo and Xiaoyi Zhang for helpful discussions. 

\bibliographystyle{siam}
\bibliography{ref_AHiSDSM}
\end{document}